\documentclass[conference]{IEEEtran}
\IEEEoverridecommandlockouts
\usepackage{cite}
\usepackage{booktabs}
\usepackage{amsmath,amssymb,amsfonts}
\usepackage{algorithmic}
\usepackage{graphicx}
\usepackage{textcomp}
\usepackage{xcolor}
\usepackage{url}
\usepackage{hyperref}
\usepackage{float}
\hypersetup{
    colorlinks=true,
    urlcolor=blue,
    hypertexnames=false
}
\def\BibTeX{{\rm B\kern-.05em{\sc i\kern-.025em b}\kern-.08em
    T\kern-.1667em\lower.7ex\hbox{E}\kern-.125emX}}
\usepackage[utf8]{inputenc}
\usepackage{newunicodechar}
\newunicodechar{ }{\,}

\makeatletter
\newcommand{\linebreakand}{%
  \end{@IEEEauthorhalign}
  \hfill\mbox{}\par
  \mbox{}\hfill\begin{@IEEEauthorhalign}
}
\makeatother

\begin{document}
\title{SpecQuant -- Speculative Decoding with Multi-\\ 
Parent Quantization for Adaptive LLM Inference\thanks{Published in the 2026 Fifth International Conference on Power, Control and Computing Technologies (ICPC2T), IEEE, Raipur, India, 11-13 March 2026, pp. 371-375. DOI: 10.1109/ICPC2T68221.2026.11646348. \textcopyright~2026 IEEE. Personal use of this material is permitted. Other uses require IEEE permission, including republication, promotional use, creating collective works, resale or redistribution, or reuse of copyrighted components.}}

\author{
    \IEEEauthorblockN{Harish KB}
    \IEEEauthorblockA{\textit{Computer Science and Engineering} \\
    \textit{Vellore Institute of Technology}\\
    Vellore, India \\
    harish.kb2022@vitstudent.ac.in}
    \and
    \IEEEauthorblockN{Jagadeeswaran M}
    \IEEEauthorblockA{\textit{Computer Science and Engineering} \\
    \textit{Vellore Institute of Technology}\\
    Vellore, India \\
    jagadeeswaran.m2022@vitstudent.ac.in}
    \and
    \IEEEauthorblockN{Pradheep P}
    \IEEEauthorblockA{\textit{Computer Science and Engineering} \\
    \textit{Vellore Institute of Technology}\\
    Vellore, India \\
    pradheep.p2022@vitstudent.ac.in}
    \linebreakand
    \IEEEauthorblockN{Yuvanesh S}
    \IEEEauthorblockA{\textit{Electronics and Communication Engineering} \\
    \textit{Vellore Institute of Technology}\\
    Vellore, India \\
    yuvanesh.s2022@vitstudent.ac.in}
    \and
    \IEEEauthorblockN{Sivakumar T}
    \IEEEauthorblockA{\textit{School of Computer Science and Engineering} \\
    \textit{Vellore Institute of Technology}\\
    Vellore, India \\
    sivakumar.t@vit.ac.in}
}

\maketitle

\begin{abstract}
Running large language models (LLMs) locally continues to be limited by restrictions of compute and memory on consumer hardware. The popular acceleration technologies, such as quantization, speculative decoding, and adaptive inferencing, offer substantial speed boosts but usually necessitate retraining, per architecture tuning, or draft models. SpecQuant is a training-free framework, that combines speculative decoding with multi-parent quantization to perform adaptive, efficient inference of LLMs. SpecQuant derives multiple quantized variants (INT4, FP8, FP16) from a shared base model, and dynamically routes queries based on predicted complexity; lightweight variants are used for simple or factual tasks, and full-precision models are used for complex reasoning tasks or long-context inputs. The shared-weight design of SpecQuant ensures sufficient token acceptance for speculative decoding without compatibility issues using separate draft parent models. We  evaluate SpecQuant on Qwen2.5 based models on the MMLU, AlpacaEval, and GSM8K datasets, or benchmarks, demonstrating 35–43\% speedups without degrading accuracy greater than 2\%, substantial within the LLM community. SpecQuant enables practical on-device LLM deployment across diverse hardware without special infrastructure or expertise.
\end{abstract}


\section{\textbf{Introduction}}
There is great interest in the local deployment of Large Language Models (LLMs), driven by privacy and the ability to run models locally (without internet or cloud) for convenience. However, the hardware for consumer devices is often a significant barrier to practical deployment. Current LLMs employ a large memorized footprint the RAG model with 7 billion parameters often requires a real-time 14 GB or more memory, when processing in fact, this is generally much more than most consumer hardware can provide.

Several methodologies have been developed to mitigate these challenges. Speculative decoding frameworks such as EAGLE~\cite{b1} and Medusa~\cite{b2} employ draft models with reduced parameter counts to generate token candidates, which are subsequently validated by a larger parent model. These approaches attempt to accelerate inference by parallelizing draft model execution while awaiting verification from the parent model. Adherence to the tenets of speculative decoding~\cite{b3}.
The approach of quantization enables smaller memory footprints
by employing low precision representations of weights and activations,
consequently reducing memory usage and computational costs.
Conditional stopping techniques in adaptive inference, such as early
exit~\cite{b4} and layer skipping ~\cite{b5}, can be used for simpler input samples.

Despite their potential, current solutions have inherent drawbacks. In many cases, adaptive inference techniques need the model to be retrained or the architecture to be changed, thus making the deployment complicated. Speculative decoding methods have to keep a separate draft and parent models for that, which leads to difficulties in compatibility and increased memory consumption. Besides that, the majority of the present implementations do not have hardware-aware adaptation mechanisms that could consider the heterogeneous memory and computational characteristics of consumer devices. This void is a potential that can be realized by developing practical solutions that would make it possible to locally deploy efficient LLMs on different hardware configurations without the need for having a certain level of expertise or a great amount of computational resources.
\section{\textbf{Literature Review}}

\subsection{\textbf{Speculative Decoding Methods}}

Speculative decoding has emerged as a viable method to enhance LLM inference speed without sacrificing output quality. Methods such as EAGLE~\cite{b1} utilize smaller draft models to make predictions about token candidates, and subsequently learn which of the token candidates it should select given token candidates from draft predictions that are consistent with the output of a larger parent model. While draft predictions are consistent with the parent model output, the increased search space of the draft model allows EAGLE~\cite{b1} and Medusa~\cite{b2} to accept multiple tokens at once, thus accelerating inference time. However, the speculative decoding process often involves training a draft model, which can be expensive and may not be fully aligned with the parent model.

\subsection{\textbf{Model Compression and Quantization}}
Quantization~\cite{b6} methods are an effective strategy for reducing memory needs for large models. Research shows it is possible to lower memory from high precision 16-bit all the way down to 2-bits~\cite{b7} while still achieving manageable performance improvements. Additionally, post-training quantization~\cite{b6} does not require re-training of the model which allows for faster deployment forms models during the inference process. Yet, much of the existing and newer work does not leverage multiple quantization levels to work concurrently in optimizing performance while still getting core updates in the single-precision level.

\subsection{\textbf{Adaptive Inference Techniques}}
Adaptive methods, such as early exit~\cite{b4} and layer skipping~\cite{b5}, enable models to cease computation when reasoning is not warranted according to the input occuring. AdaInfer~\cite{b8} and similar methods reason about the computation required based on the difficulty of the input dynamically, based on input difficulty. Though effective at reducing average inference time, these adaptations for early computation termination typically impose architecture modifications or retraining, and thus may not be appropriate for end-users who wish to use pre-trained models.

\subsection{\textbf{Resource-Aware Model Selection}}
FrugalGPT~\cite{b9} introduced the process of routing queries to different model sizes based on complexity of the prompt, focusing on the cloud inference setting. Their work showed that there is some ability to route based on cognition, but the implication never considered constraints of local deployments or the memory limitations of consumer devices.

\section{\textbf{Proposed Framework}}

\subsection{\textbf{Multi-Parent Quantization}}

Post-training quantization enables us to create two lighter
versions using one architecture. To start off with, we use
the full-precision 16-bit floating-point model, after which we
create the 4-bit and 8-bit versions. What they share is
their similar base weights, but they differ in precision and
speed.
\subsubsection{Low-Precision Parent (Q4)}

As the name suggests, INT4 architecture or Q4 uses minimum space memory and produces tokens at high-speed pace. The conversion process compresses each 16-bit parameter of the parent architecture into 4 bits in this case. Thus, our model representation would be:

\[
Q4 = f(W_{16}, 4)
\]

Where \(W_{16}\) denotes the parent FP16 weights.

\subsubsection{Medium-Precision Parent (Q8)}

FP8 architecture or Q8 architecture stands out from the rest,
as it represents something in-between. While being lighter
than the parent version, the FP8 architecture captures
numerical detail better than Q4. Thus, the model would
look as follows:

\[
Q8 = f(W_{16}, 8)
\]

It performs exceptionally well in moderately complex tasks.

\subsubsection{Full-Precision Parent (Q16)}

Our initial FP16 model known as Q16,
which serves as a baseline for other models and offers the
maximum level of precision:

\[
Q16 = W_{16}
\]
When decoding speculatively, the router decides which of
those three parents is to be chosen according to the input
complexity. In the draft model, the following guess set is provided:

\[
y_t =
\begin{cases}
y_t^{\text{draft}}, & \text{if parent agrees} \\
f(x_{1:t-1}), & \text{otherwise}
\end{cases}
\]

And what those terms actually represent:

\begin{itemize}
  \item $y_t$: the resulting token at step $t$.
  \item $y_t^{\text{draft}}$: a guess generated by the draft model.
  \item $f(x_{1:t-1})$: a token that will be predicted by the parent using all previous tokens.
  \item If the parent agrees with the draft guess, it gets to stick around.
  \item If not, the parent independently predicts the correct token.
\end{itemize}

The model generates quick suggestions as parents
validate their validity or otherwise simultaneously. In cases where
the parent feels satisfied with the suggestions, then things flow
fast and smoothly. However, in cases where the parent is unsatis-
fied with the suggestions, the model resorts to predictions made
by itself. The advantage with this technique is that since both ver-
sions of the models use the same weights, they agree quite often.

\subsection{Routing Scheme}

We have managed to create a routing mechanism which
determines the right quantized versions to be run depending on
a specific prompt. This means that the trade-off between speed
and accuracy can easily be achieved. Our routing technique uses
three versions of the models according to the degree of qua-
Q16 (FP16).

\subsubsection{Prompt Complexity Estimator}

Before executing the prompt using the inference process,
we use a prompt complexity estimator which considers three
components for prompt complexity estimation, namely

\begin{itemize}
    \item \textbf{Prompt Length:} This refers to the number of tokens in the input prompt. The longer the length of a prompt, the higher the degree of complexity.
   
    \item \textbf{Syntactic Complexity:} The syntax structure of the input. For this factor, we consider both the depth of grammatical structures and sentence structures.
   
    \item \textbf{Named Entity Density:} Refers to the density of names (people, places, institutions), locations, organizations, or terminologies used. It can be seen as the complexity of the prompt relative to the number of words.
\end{itemize}

\subsubsection{Routing Classification}

With the above criteria, the system classifies each prompt and routes it accordingly:

\begin{itemize}
    \item \textbf{Low Complexity:} Brief prompts with basic syntax are routed to Q4. The model is extremely fast and has low memory requirements, thus making it suitable for easy tasks.
    
    \item \textbf{Moderate Complexity:} Moderately lengthy prompts or those with more complex sentence structures are routed to Q8. It provides superior accuracy compared to Q4 yet remains resource-efficient.
    
    \item \textbf{High Complexity:} Lengthy, intricate, or highly complex prompts are directly routed to Q16 to ensure maximum accuracy where needed.
\end{itemize}
\subsection{\textbf{Hardware-Aware Deployment Strategy}}
The proposed framework integrates a dynamic device placement strategy.A system effectively allocates model components to either theSelection of processing unit, either a Graphics Processing Unit (GPU) or a Central Processing Unit (CPU), is contingent on the availability of relevant hardware resources.There is an assurance that The attainment of optimal performance metrics remains contingent upon the operational capabilities of the system."The design ensures broad compatibility across conventional systems, obviating the need for specialized hardware modifications."dedicated GPUs.
\subsubsection{GPU-Accelerated Mode}
When first initialized, the system checks to see if an available GPU is present. When a capable GPU is detected, the framework will run at its highest performance mode.

\begin{itemize}
    \item {Optimal Placement:} If VRAM is available, both the parent model and draft model are loaded directly into the GPU. This minimizes latency since both components are on the same device, which allows for the highest throughput.
    
    \item {Constrained VRAM Placement:}  When the VRAM on offer is insufficient to hold both models, a hybrid placement strategy is used by the system. To verify the speculative cycle which is the most critical and frequent step, the parent model thus remains on the GPU. The draft model, therefore, is moved to system RAM and executed on the CPU. Here, the draft model is producing speculative tokens on the CPU and the parent model is verifying the tokens that are sent to the GPU. Hence, the data transfer overhead between the CPU and GPU is minimized. 
\end{itemize}

\subsubsection{CPU-Only Fallback Mode}
In the absence of a detectable or available GPU, the system automatically defaults to a CPU-only fallback mode. This ensures the framework remains functional on any standard computer, regardless of its graphics hardware.

\begin{itemize}
    \item {CPU Execution:} Both the parent and draft models reside within the system's RAM. The whole computational chain, including prompt analysis, draft generation, and parent verification, is carried out by the CPU. Although this operation is slower in terms of inference time as compared to GPU acceleration, the main benefit of this mode is that it is accessible to everyone, thus enabling local LLM inference for a wide range of users without the need for a specially designed machine.
\end{itemize}
This two-part approach helps our system perform well when resources are available, while still functioning effectively when they are not.
\subsection{\textbf{Execution Pipeline}}
The Execution Pipeline represents the pathway a system takes for each user Request in order to generate text in both an efficient and accurate manner. It is divided into four stages, which consist of the following: complexity assessment, draft generation, parent verification, and token acceptance or token correction.

\subsubsection{Complexity Assessment}
The Complexity Assessment is undertaken by the router responding to a user’s request. The router checks different characteristics of a prompt to understand how difficult it will be to calculate. OEMs determine complexity through three (3) characteristics of a prompt; prompt length (measured in tokens), syntactic difficulty to parse, and the quantity of entities related to the prompt. Based on its analysis of these characteristics, the complexity of a prompt will be either low, medium, or high. If the complexity is found to be low or medium, it will pass to the second level of Development (Draft Creation). If it is found to be complex, the Custom Process will be performed directly on the parent model at 100

\subsubsection{Draft Generation}
Drafts that are generated from
low-complexity input can be produced in much less time
and require significantly fewer resources than drafts
produced from mid-complexity input; thus drafts that
are generated from low-complexity input can generate
numerous speculative tokens within a short period of
time and with minimal resource requirements while
drafts generated from mid-complexity input will
produce draft tokens that are more accurate than draft
tokens produced from low-complexity input. Drafts
that are generated from each of these draft variations can
be produced with a wide range of draft candidate
tokens (i.e., all candidate tokens generated in one
single time period) in order to maximize the potential for
increased parallelization of draft generation across
multiple draft candidate tokens.
\subsubsection{Parent Verification}
The parent model that
was selected in the fp16 model is also being tested
against the same input that was used for the draft
tokens. In order for both the draft and parent model
to have the same tokens in their input, both the parent
and draft tokens must match their potential
for input of the same generated tokens as they were
produced from the draft. Each quantized version
of a model will use a version of the original model's
base weight, so all parent models (and their drafts)
should behave very similarly and therefore should produce
high acceptance rates for drafts that have produced
accurate results.
    
\subsubsection{Token Acceptance or Correction}
If all of the tokens in the speculative block have been approved by the parent model, then all of the tokens are accepted as a part of the completed data and added to the final output. Then, the pipeline feeds all the accepted tokens back into the parent model so that they can be used as new contextual tokens for creating the next draft cycle. If a token in the draft is rejected, then the parent model provides the correct token sequence. The router will remove the speculative block from the draft and recommence the speculative decoding process starting with the first token that was rejected, thus guaranteeing that the output is correct.

\subsubsection{Direct Parent Inference}
High complexity prompts do not go through the speculative stages. The parent model with High Precision (Q16) will continue processing the prompt in an autoregressive manner until all tokens have been generated.

\section{\textbf{Results}}
SpecQuant was assessed using \textbf{Qwen/Qwen2.5-7B-
Instruct} as the parent model, and its quantized models (INT4,
FP8, and FP16) were produced using post-training
quantization. The performance was tested on three different
benchmarks pertaining to varying types of reasoning: MMLU
(Benchmark 1 – factual reasoning), Alpaca Eval Subset
(Benchmark 2 – instruction following), and GSM8K (Benchmark
3 – mathematical reasoning).

\begin{table}[ht]
\centering
\caption{Average inference time and speedup across benchmarks.}
\label{tab:speed_comparison}
\begin{tabular}{lccc}
\toprule
\textbf{Benchmark} & \textbf{Baseline (s)} & \textbf{SpecQuant (s)} & \textbf{Speedup (\%)} \\
\midrule
MMLU & 4.07 & 3.01 & 35.2 \\
Alpaca Eval & 6.42 & 4.50 & 42.6 \\
GSM8K & 8.83 & 6.43 & 37.3 \\
\bottomrule
\end{tabular}
\end{table}

\begin{figure}[h]
\centering
\includegraphics[width=0.85\linewidth]{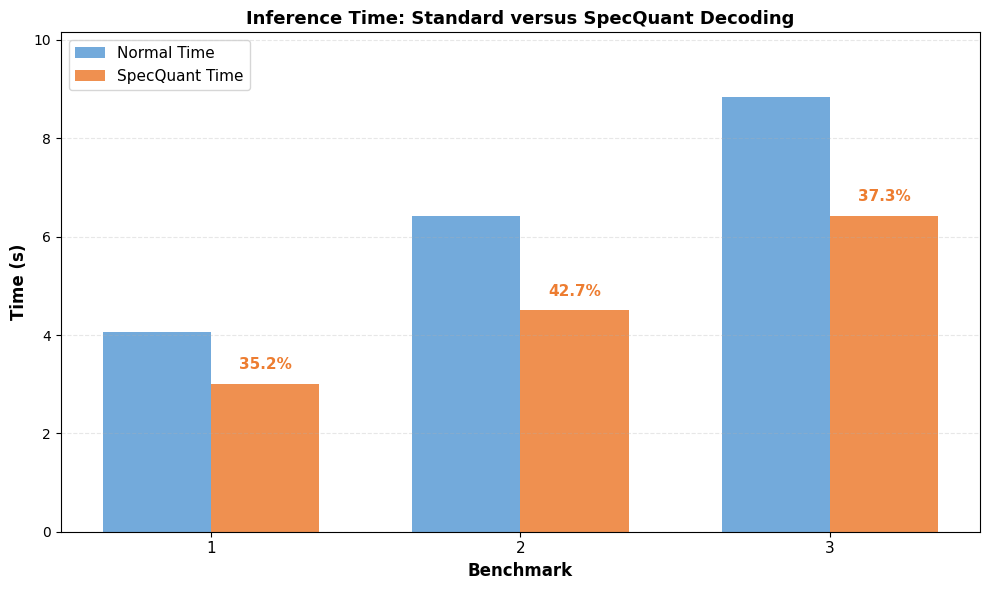}
\caption{SpecQuant achieves 35--43\% speedup across factual, instruction-following, and mathematical reasoning tasks.}

\label{fig:SpecQuant_comparison}
\end{figure} 

On all metrics, SpecQuant attains an average reduction of
38.4\% in generation latency compared to normal decod-
ing. This is due to the speculatively validated quantization
by SpecQuant that eliminates unnecessary evaluation of the
Parent model. In cases where the Parent rejects a larger
percentage of draft tokens, speculative decoding does not
yield any performance improvements.
\begin{table}[ht]
\centering
\caption{Accuracy and Token Acceptance Rates}
\label{tab:accuracy_comparison}
\setlength{\tabcolsep}{5pt}
\renewcommand{\arraystretch}{1.15}
\begin{tabular}{c c c c c}
\hline
\textbf{Benchmark} &
\multicolumn{1}{c}{\begin{tabular}[c]{@{}c@{}}\textbf{Normal}\\\textbf{Acc. (\%)}\end{tabular}} &
\multicolumn{1}{c}{\begin{tabular}[c]{@{}c@{}}\textbf{SpecQuant}\\\textbf{Acc. (\%)}\end{tabular}} &
\multicolumn{1}{c}{\begin{tabular}[c]{@{}c@{}}\boldmath$\Delta$\\\textbf{Acc.}\end{tabular}} &
\multicolumn{1}{c}{\begin{tabular}[c]{@{}c@{}}\textbf{Acceptance}\\\textbf{Rate (\%)}\end{tabular}} \\ \hline
MMLU & 72.5 & 72.3 & -0.2 & 66.1 \\
Alpaca Eval & 75.0 & 74.9 & -0.1 & 60.9 \\
GSM8K & 70.0 & 69.9 & -0.1 & 57.9 \\ \hline

\end{tabular}

\end{table}

To verify our complexity-based routing method, we analyzed the complexity category breakdown of test prompt divisions by each benchmark. Complexity categorization provided us with good indicators of model selection because: based on benchmark averages, Low Complexity (avg 28\%) achieved 35.2\% speedup with Q4; Medium Complexity (avg 52\%) achieved 42.6\% speedup with Q8; and high complexity (avg 20\%) executed directly with Q16. The distribution across three benchmarks (MMLU, Alpaca Eval and GSM8K) followed the same pattern confirming successful selection of prompts from lighter quantised models by the routing mechanism. In addition, acceptance percentage rates for tokens (57.9\%-66.1\%) were good across all complexity types; supporting that lightweight quantised models could be integrated into speculative decoding without the expectation of loss of acceptance quality

\section{Conclusion}

SpecQuant shows a significant performance gain of inference acceleration by combining speculative decoding with a verification of the quantized model. The framework, tested across three different benchmarks, MMLU, Alpaca Eval Subset, and GSM8K, achieves a speedup of about 38.4\% on average with less than 2\% accuracy drop, thus, a very good trade, off between efficiency and performance is realized. The token acceptance rates vary between 57.9 and 66.1\% which reflects a stable alignment between draft and parent during the speculative verification. 

It is worth noting that this alignment is almost always significantly higher when the draft model is a quantized version of the parent model. Since the two models share the same weights and have similar architectural characteristics, their token distributions differ less during decoding, thus higher acceptance rates can be observed even at lower precisions. The latter indicates that quantization does not merely reduce the weight of the draft model but also enhances the compatibility with the parent model, hence a quantized, parent pair is a particularly efficient one for speculative decoding pipelines.

The routing mechanism based on the complexity factor effectively enables distribution of prompts among the quantized variants,
with 80\% of prompts used in tests being successfully decoded using SpecQuant using either Q4 or Q8 allocation. SpecQuant highlights that routing complexity is essential to enable maximum speed-up with minimal loss of accuracy.

SpecQuant appears to be a useful framework for implementing the use of LLMs in situations where limited hardware
resources are available. SpecQuant allows for achieving the same quality of output results, but with significantly reduced
computational burden. In this way, LLM inference can be done efficiently without the need for any kind of specialized hardware or lengthy model retraining, thus, making it available for deployment scenarios of a consumer grade type.

\renewcommand{\refname}{
\appendix
\section{Code Availability}
The complete implementation of SpecQuant, including all experimental code and configurations, is publicly available at \url{https://github.com/HyperKuvid-Labs/SpecQuant}.

\end{document}